\documentclass[letterpaper, 10 pt, conference]{ieeeconf}  

\usepackage{graphicx}
\usepackage{booktabs}
\usepackage{times}
\usepackage{soul}
\usepackage{color}
\usepackage{xcolor}
\usepackage{url}
\usepackage{amsfonts,amssymb}
\usepackage[hidelinks]{hyperref}
\usepackage[utf8]{inputenc}
\usepackage[small]{caption}
\usepackage{graphicx}
\usepackage{amsmath}

\usepackage{amsthm}

\usepackage{booktabs}
\usepackage{algorithm}
\usepackage{xcolor} 
\usepackage{algorithmic}
\usepackage[switch]{lineno}
\usepackage{booktabs}
 \usepackage{multirow}
 \usepackage{pifont}
 \usepackage{bm}
 \usepackage{tabularx}
 \usepackage{adjustbox}
 \usepackage{makecell}
\usepackage{array}  
\usepackage{caption}

\newcommand{\dataset}[0]{RAD}
\IEEEoverridecommandlockouts                              

\title{\LARGE \bf
\dataset{}: A Comprehensive Dataset for Benchmarking the Robustness of Image Anomaly Detection
}

\author{Yuqi~Cheng, Yunkang~Cao, Rui~Chen, Weiming~Shen, \textit{Fellow, IEEE}
\thanks{Resrach supported by the Fundamental Research Funds for the Central Universities: HUST:2021GCRC058.
}
\thanks{Yuqi Cheng, Yunkang Cao, Rui Chen, and Weiming Shen are with State Key Laboratory of Intelligence Manufacturing Equipment and Technology, School of Mechanical Science and Engineering, Huazhong University of Science and Technology, Wuhan, China. Their emails are yuqicheng@hust.edu.cn, cyk\_hust@hust.edu.cn, 591609840@qq.com, wshen@ieee.org (corresponding author)
}}

\usepackage{cite}
\begin{document}

\maketitle
\thispagestyle{empty}
\pagestyle{empty}

\begin{abstract}

Robustness against noisy imaging is crucial for practical image anomaly detection systems. This study introduces a Robust Anomaly Detection (RAD) dataset with free views, uneven illuminations, and blurry collections to systematically evaluate the robustness of current anomaly detection methods. Specifically, \dataset{} aims to identify foreign objects on working platforms as anomalies. The collection process incorporates various sources of imaging noise, such as viewpoint changes, uneven illuminations, and blurry collections, to replicate real-world inspection scenarios. Subsequently, we assess and analyze 11 state-of-the-art unsupervised and zero-shot methods on \dataset{}. Our findings indicate that: 1) Variations in viewpoint, illumination, and blurring affect anomaly detection methods to varying degrees; 2) Methods relying on memory banks and assisted by synthetic anomalies demonstrate stronger robustness; 3) Effectively leveraging the general knowledge of foundational models is a promising avenue for enhancing the robustness of anomaly detection methods. The dataset is
available at https://github.com/hustCYQ/RAD-dataset.

\end{abstract}


\section{INTRODUCTION}

Anomaly detection aims to identify deviations from the normal pattern distribution~\cite{cao2024survey}. It plays a crucial role in intelligent manufacturing systems by detecting abnormal behaviors in industrial scenarios and surface defects in industrial products.

The complexity of detection environments and the variety of objects in industrial settings pose significant challenges to the robustness of anomaly detection methods. Benefiting from recent anomaly detection datasets like RealIAD~\cite{wang2024real}, VisA~\cite{VisA}, and MVTec~\cite{mvtec}, the development of anomaly detection methods~\cite{zhang2023exploring, yao2023scalable} has been greatly promoted. However, evaluations of anomaly detection methods predominantly rely on aligned datasets with high-quality imaging~\cite{xie2023iad, akcay2022anomalib}, a practice that may not reflect real-world conditions. In practical industrial scenarios, image quality may be compromised due to variations in camera viewpoints~\cite{cheng2022novel}, uneven environmental brightness in the environment~\cite{cheng2021novel}, and collection noises, necessitating robust anomaly detection methods. To better replicate practical detection scenarios, some datasets introduce imaging noise. For instance, MPDD~\cite{MPDD} captures images of objects from multiple angles through rotation but vertically aligns the objects in images, deviating from reality. Additionally, the EyeCandies~\cite{eyecandies} dataset employs multiple light sources, yet the placement of these sources remains constant, limiting the variability of lighting conditions. Despite these datasets considering various imaging factors, they do not thoroughly benchmark the robustness of anomaly detection methods against the complexity of practical scenarios, which may include diverse imaging noises. Consequently, there is an urgent need for conducting robustness evaluations of anomaly detection methods with data that closely resembles actual industrial scenarios, benefiting both methodological research and practical applications.

\begin{figure}[t!]
\centering\includegraphics[width=\linewidth]{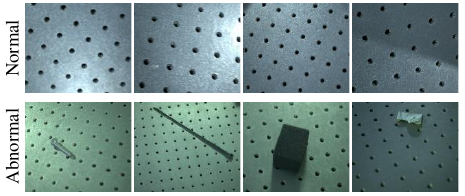}
\caption{Samples from \dataset{} dataset. The first row contains some normal samples. The second row contains some abnormal samples and the categories of foreign objects from left to right are bolt, cable tie, sponge, and tape.}
\vspace{-0.5cm}
\label{fig:data}
\end{figure}

To comprehensively benchmark the robustness of anomaly detection methods, we introduce a Robust Anomaly Detection (RAD) dataset, designed to detect foreign objects under conditions such as free camera views, uneven illuminations, and blurry image collections, thereby advancing the evaluation of anomaly detection method robustness. Contrary to traditional benchmarks that focus on identifying defects in abnormal products, our dataset is constructed in a straightforward yet highly effective manner. Specifically, we consider a clean work platform in practical industrial production as the normal state. Instances of machine wear or human error may result in foreign objects falling onto the platform, thereby disrupting operations and posing safety hazards. Consequently, detecting foreign objects on the work platform is framed as an anomaly detection challenge, for which we design scenarios representing both normal and abnormal conditions. 

As illustrated in Fig.~\ref{fig:data}, normal scenarios depict a work platform with multiple metal holes, whereas the abnormal scenarios feature the work platform with foreign objects. To establish a comprehensive benchmark for anomaly detection robustness, we introduce different types of foreign objects onto the work platform, including bolts, cable ties, sponges, and tapes. Subsequently, the images are captured under various imaging noises, encompassing random viewpoints, illuminations, and imaging distances to replicate practical noise scenarios. Then we systematically benchmark the robustness of state-of-the-art (SOTA) anomaly detection methods on the established dataset, offering valuable insights into the pathway to robust anomaly detection methods.

The contributions of this study are summarized as follows:

\begin{itemize}
\item We propose the \dataset{} dataset, characterized by free views, uneven illuminations, and blurry collections, for the robustness evaluation of anomaly detection.
\item Extensive experiments are conducted to evaluate the performance of SOTA anomaly detection methods, thereby establishing a comprehensive benchmark for research into anomaly detection robustness.
\item The robustness of various anomaly detection methods is analyzed under conditions of free views, uneven illuminations, and blurry collections, highlighting potential research directions for enhancing anomaly detection robustness.
\end{itemize}

The remainder of the paper is organized as follows: Section \ref{sec:dataset} details the construction process and characteristics of the \dataset{} dataset. Section \ref{sec:exp} benchmarks existing mainstream methods on the \dataset{} dataset. Finally, the conclusion is presented in Section \ref{sec:conclusion}.

\section{DATASET}\label{sec:dataset}
\subsection{Data collecting}

\begin{figure}[h]
\centering\includegraphics[width=\linewidth]{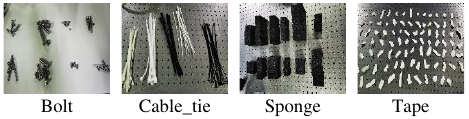}
\caption{Foreign objects used in data collection: bolt, cable tie, sponge, and tape.}
\label{fig:object}
\end{figure}

The \dataset{} dataset is constructed by Basler acA2440 industrial camera with a resolution of 2440×2048 to collect images of work platform with/without foreign objects. The work platform consists of a metal plate with multiple holes. As shown in Fig.~\ref{fig:object}, there are foreign objects prepared in advance, including four categories: Bolt, Cable\_tie, Sponge, and Tape:

\begin{itemize}
\item The Bolt category comprises bolts and nuts in various sizes and diameters;
\item The Cable\_tie category encompasses cable ties of different sizes, widths, and black and white colors.
\item The Sponge category consists of sponge cubes cut into different sizes.
\item The tape category is manually created to simulate the actual used tapes.
\end{itemize}

\begin{table}[h]
\centering
\caption{The number of training and testing samples for each category in the \dataset{} dataset.}
\label{table:number}
\setlength{\tabcolsep}{18.pt}
\resizebox{\linewidth}{!}
{
\begin{tabular}{@{}c|c|c|c@{}}
\toprule[1.5pt]
\multirow{2}{*}{Category} & \multirow{2}{*}{Train Samples} & \multicolumn{2}{c}{Test   Samples} \\ \cmidrule(l){3-4} 
          &  & Normal & Abnormal \\ \midrule
Bolt                      & \multirow{4}{*}{213}           & \multirow{4}{*}{73}      & 327     \\ 
Cable\_tie &  &        & 293      \\ 
Sponge    &  &        & 281      \\ 
Tape      &  &        & 323      \\ \bottomrule[1.5pt]
\end{tabular}
}
\end{table}
\begin{figure}[h]
\centering\includegraphics[width=\linewidth]{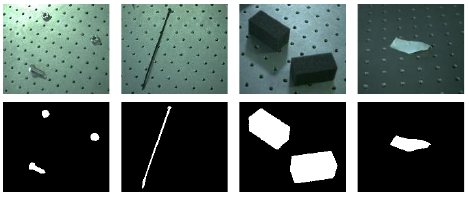}
\caption{Original images and corresponding pixel-wise masks from \dataset{} dataset.}
\label{fig:mask}
\vspace{-0.5cm}
\end{figure}
\begin{figure*}[t!]
\centering\includegraphics[width=\linewidth]{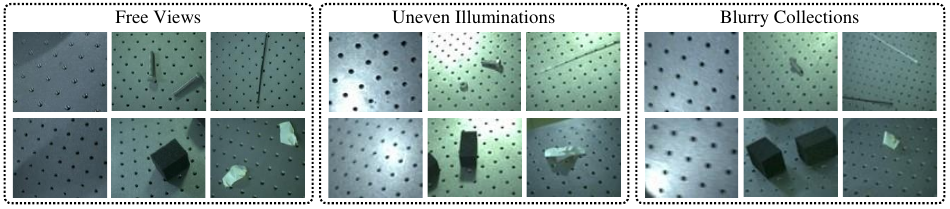}
\caption{Samples from \dataset{} dataset. (Left) Foreign object images collected from different perspectives. (Middle) Foreign object images with uneven illumination. (Right) Foreign object images with blurry.}
\label{fig:images}
\vspace{-0.5cm}
\end{figure*}



Given the diverse imaging noises prevalent in practical applications, our approach takes into account viewpoints, illuminations, and blurs during the construction of \dataset{} and proposes three types of noises, \textit{i.e.}, free views, uneven illuminations, and blurry collections. 

\subsubsection{Free View}

Free views involve collecting images from multiple angles, where foreign objects are not necessarily centered in the field of view, as shown in Fig.~\ref{fig:images} (Left). Unlike other datasets, foreign objects are not registered to similar positions in this dataset, which results in inconsistent features at the same position in different images, requiring the model to have stronger global feature processing capabilities. In addition, although the holes in the background are similar, their distribution in different images are biased due to lack of alignment, which increases the requirement for the background understand of models.

\subsubsection{Uneven Illuminations}

Uneven illuminations occur when object areas become excessively bright, creating bright spots, as shown in Fig.~\ref{fig:images} (Middle). The shape of bright spots are different, and bright spots cause deviations in the pixels/features distribution of images, which may lead to the models erroneously identifying the bright spots as anomalies.
In addition, uneven illuminations can also create strong shadows around foreign objects, which is not actually an anomaly, but may prompt models to misidentify shadows as part of the foreign object.

\subsubsection{Blurry Collections}

By adjusting the acquisition distance of the camera or reducing its stability, images may exhibit out-of-focus blurs and motion blurs, as shown in Fig.~\ref{fig:images} (Right). Blurs can reduce imaging accuracy and make the boundaries of foreign objects inaccurate, necessitating models to more precisely recognize object contours. Furthermore, as a form of noise, blurs can also disrupt the model's ability to recognize features such as holes in the background.

Within the dataset, the aforementioned characteristics are interlinked, with the collected images often exhibiting two or three of these features simultaneously. Consequently, this complex dataset can facilitate the exploration of robustness in anomaly detection methods.



During the collection process, multiple images of the work platform without foreign objects are captured to serve as normal samples. Subsequently, prepared foreign objects are placed on the work platform for the collection of abnormal samples. The distribution of samples in each category is shown in Table~\ref{table:number}. Following the mainstream unsupervised anomaly detection setting~\cite{mvtec}, we collected a total of 286 normal samples, of which 213 are used for training, and the remaining 73 for testing. Additionally, 327, 293, 281, and 323 abnormal samples across the four categories are used for testing, respectively. The size of the images is standardized to 612×512 pixels. Each abnormal sample has been accurately labeled, as depicted in Fig.~\ref{fig:mask}.

\section{EXPERIMENT}\label{sec:exp}

\begin{table*}[t!]
\centering
\caption{Quantitative image-level Results On \dataset{} Dataset. The Best Is In Bold, And The Second Best In Underlined. (\%)}
\label{table:image-level}
\setlength{\tabcolsep}{4.pt}
\resizebox{\linewidth}{!}
{
\begin{tabular}{@{}ccccccccccccc@{}}
\toprule[1.5pt]
\multirow{2}{*}{Category} &
  \multirow{2}{*}{Metric} &
  \multicolumn{2}{c}{Flow-based} &
  \multicolumn{4}{c}{Knowledge-Distillation-based} &
  \multicolumn{2}{c}{Memory-Bank-based} &
  \multicolumn{3}{c}{Zero-Shot} \\ \cmidrule(l){3-4} \cmidrule(l){5-8} \cmidrule(l){9-10} \cmidrule(l){11-13}
 &
   &
  Cflow~\cite{cflow} &
  PyramidFlow~\cite{PyramidFlow} &
  RD~\cite{RD} &
  RD++~\cite{RD++} &
  DSR~\cite{DSR} &
  CDO~\cite{cdo} &
  Patchcore~\cite{patchcore} &
  GCPF~\cite{GCPF} &
  SAA~\cite{saa} &
  WinCLIP~\cite{winclip} &
  APRIL-GAN~\cite{APRIL-GAN} \\ \midrule
\multirow{3}{*}{Bolt} &
  AUROC &
  92.0 &
  90.9 &
  95.4 &
  \textbf{97.9} &
  {\underline{96.4}} &
  84.0 &
  93.7 &
  92.5 &
  75.3 &
  96.1 &
  88.8 \\
 &
  AP &
  98.0 &
  97.5 &
  98.8 &
  \textbf{99.4} &
  {\underline{99.3}} &
  92.9 &
  98.2 &
  98.0 &
  94.4 &
  99.1 &
  97.4 \\
 &
  Max-F1 &
  93.1 &
  93.2 &
  95.5 &
  \textbf{97.9} &
  {\underline{95.4}} &
  91.7 &
  95.1 &
  94.2 &
  89.8 &
  95.0 &
  92.1 \\ \midrule
\multirow{3}{*}{Cable\_tie} &
  AUROC &
  \textbf{98.9} &
  94.1 &
  98.1 &
  97.5 &
  93.9 &
  93.6 &
  {\underline{98.7}} &
  97.0 &
  20.0 &
  98.6 &
  88.1 \\
 &
  AP &
  \textbf{99.7} &
  98.2 &
  99.3 &
  99.2 &
  98.3 &
  95.0 &
  99.5 &
  98.7 &
  70.2 &
  \textbf{99.7} &
  96.4 \\
 &
  Max-F1 &
  99.2 &
  95.2 &
  \textbf{99.3} &
  97.5 &
  93.9 &
  97.8 &
  \textbf{99.3} &
  99.0 &
  89.1 &
  97.8 &
  92.2 \\ \midrule
\multirow{3}{*}{Sponge} &
  AUROC &
  {\underline{99.6}} &
  82.8 &
  97.4 &
  97.9 &
  84.6 &
  96.4 &
  98.5 &
  95.0 &
  92.2 &
  \textbf{99.8} &
  57.2 \\
 &
  AP &
  {\underline{99.9}} &
  94.7 &
  98.8 &
  99.1 &
  95.9 &
  97.0 &
  99.3 &
  96.8 &
  97.9 &
  \textbf{100.0} &
  86.2 \\
 &
  Max-F1 &
  99.1 &
  89.7 &
  \textbf{99.3} &
  99.1 &
  90.9 &
  99.1 &
  \textbf{99.3} &
  98.6 &
  93.8 &
  \textbf{99.3} &
  88.5 \\ \midrule
\multirow{3}{*}{Tape} &
  AUROC &
  {\underline{99.9}} &
  96.3 &
  99.3 &
  {\underline{99.9}} &
  99.7 &
  97.2 &
  99.2 &
  98.2 &
  44.6 &
  \textbf{100.0} &
  97.3 \\
 &
  AP &
  \textbf{100.0} &
  99.1 &
  99.8 &
  \textbf{100.0} &
  99.9 &
  97.6 &
  99.7 &
  99.4 &
  85.5 &
  \textbf{100.0} &
  99.4 \\
 &
  Max-F1 &
  {\underline{99.7}} &
  96.0 &
  99.5 &
  {\underline{99.7}} &
  99.1 &
  99.4 &
  99.5 &
  99.5 &
  89.8 &
  \textbf{99.8} &
  96.0 \\ \midrule
\multirow{3}{*}{Mean} &
  AUROC &
  97.6 &
  91.0 &
  97.6 &
  {\underline{98.3}} &
  93.7 &
  92.8 &
  97.5 &
  95.7 &
  58.0 &
  \textbf{98.6} &
  82.9 \\
 &
  AP &
  {\underline{99.4}} &
  97.4 &
  99.2 &
  {\underline{99.4}} &
  98.4 &
  95.6 &
  99.2 &
  98.2 &
  87.0 &
  \textbf{99.7} &
  94.9 \\
 &
  Max-F1 &
  97.8 &
  93.5 &
  {\underline{98.4}} &
  \textbf{98.6} &
  94.8 &
  97.0 &
  98.3 &
  97.8 &
  90.6 &
  98.0 &
  92.2 \\ \bottomrule[1.5pt]
\end{tabular}
}
\end{table*}
\begin{table*}[t!]
\centering
\caption{Quantitative pixel-level Results On \dataset{} Dataset. The Best Is In Bold, And The Second Best In Underlined. (\%)}
\label{table:pixel-level}
\setlength{\tabcolsep}{4.pt}
\resizebox{\linewidth}{!}
{
\begin{tabular}{@{}ccccccccccccc@{}}
\toprule[1.5pt]
\multirow{2}{*}{Category} &
  \multirow{2}{*}{Metric} &
  \multicolumn{2}{c}{Flow-based} &
  \multicolumn{4}{c}{Knowledge-Distillation-based} &
  \multicolumn{2}{c}{Memory-Bank-based} &
  \multicolumn{3}{c}{Zero-Shot} \\ \cmidrule(l){3-4} \cmidrule(l){5-8} \cmidrule(l){9-10} \cmidrule(l){11-13}
 &
   &
  Cflow~\cite{cflow} &
  PyramidFlow~\cite{PyramidFlow} &
  RD~\cite{RD} &
  RD++~\cite{RD++} &
  DSR~\cite{DSR} &
  CDO~\cite{cdo} &
  Patchcore~\cite{patchcore} &
  GCPF~\cite{GCPF} &
  SAA~\cite{saa} &
  WinCLIP~\cite{winclip} &
  APRIL-GAN~\cite{APRIL-GAN} \\ \midrule
\multirow{4}{*}{Bolt} &
  AUROC &
  96.2 &
  92.4 &
  \textbf{99.1} &
  98.5 &
  95.6 &
  {\underline{98.9}} &
  97.8 &
  98.0 &
  54.7 &
  81.7 &
  98.7 \\
 &
  AP &
  29.5 &
  8.4 &
  \textbf{54.5} &
  {\underline{49.4}} &
  38.4 &
  44.7 &
  46.3 &
  45.5 &
  1.2 &
  9.1 &
  47.7 \\
 &
  Max-F1 &
  35.5 &
  16.6 &
  \textbf{55.8} &
  {\underline{51.8}} &
  44.7 &
  51.2 &
  49.4 &
  48.8 &
  2.5 &
  15.9 &
  50.2 \\
 &
  AUPRO &
  79.8 &
  75.4 &
  \textbf{95.3} &
  92.6 &
  87.6 &
  {\underline{94.1}} &
  85.4 &
  89.2 &
  18.5 &
  42.5 &
  85.1 \\ \midrule
\multirow{4}{*}{Cable\_tie} &
  AUROC &
  93.4 &
  87.1 &
  95.2 &
  95.4 &
  94.7 &
  \textbf{97.6} &
  90.9 &
  {\underline{96.5}} &
  42.9 &
  77.1 &
  95.2 \\
 &
  AP &
  29.8 &
  10.9 &
  29.2 &
  28.4 &
  46.4 &
  \textbf{40.4} &
  16.1 &
  {\underline{37.3}} &
  1.9 &
  6.2 &
  30.7 \\
 &
  Max-F1 &
  36.6 &
  18.3 &
  36.6 &
  36.9 &
  \textbf{48.9} &
  {\underline{48.8}} &
  27.8 &
  44.0 &
  4.5 &
  11.8 &
  39.4 \\
 &
  AUPRO &
  69.5 &
  56.6 &
  84.7 &
  \textbf{95.6} &
  81.3 &
  {\underline{90.4}} &
  63.3 &
  88.8 &
  11.9 &
  41.5 &
  63.5 \\ \midrule
\multirow{4}{*}{Sponge} &
  AUROC &
  93.4 &
  49.6 &
  {\underline{95.0}} &
  86.6 &
  92.2 &
  \textbf{99.0} &
  92.4 &
  88.8 &
  60.5 &
  70.7 &
  90.1 \\
 &
  AP &
  55.1 &
  8.8 &
  54.9 &
  32.4 &
  {\underline{71.}1} &
  \textbf{87.5} &
  61.5 &
  46.1 &
  12.2 &
  22.5 &
  49.8 \\
 &
  Max-F1 &
  57.7 &
  18.7 &
  62.6 &
  43.4 &
  {\underline{67.5}} &
  \textbf{84.2} &
  62.8 &
  55.3 &
  21.1 &
  27.8 &
  51.9 \\
 &
  AUPRO &
  71.9 &
  8.7 &
  {\underline{84.5}} &
  61.9 &
  82.8 &
  \textbf{95.9} &
  75.0 &
  69.7 &
  22.6 &
  35.9 &
  60.3 \\ \midrule
\multirow{4}{*}{Tape} &
  AUROC &
  98.5 &
  89.7 &
  99.0 &
  98.4 &
  {\underline{99.2}} &
  {\underline{99.2}} &
  98.9 &
  99.1 &
  50.6 &
  96.9 &
  \textbf{99.3} \\
 &
  AP &
  62.1 &
  31.5 &
  67.7 &
  58.9 &
  \textbf{89.8} &
  73.6 &
  79.1 &
  {\underline{82.4}} &
  3.9 &
  58.3 &
  81.1 \\
 &
  Max-F1 &
  66.8 &
  40.4 &
  71.9 &
  64.9 &
  \textbf{83.9} &
  74.2 &
  76.7 &
  {\underline{79.1}} &
  7.7 &
  58.5 &
  78.1 \\
 &
  AUPRO &
  93.9 &
  72.1 &
  96.7 &
  94.4 &
  \textbf{97.4} &
  97.1 &
  95.5 &
  {\underline{97.2}} &
  17.7 &
  87.1 &
  95.8 \\ \midrule
\multirow{4}{*}{Mean} &
  AUROC &
  95.4 &
  79.7 &
  {\underline{97.1}} &
  94.7 &
  95.4 &
  \textbf{98.7} &
  95.0 &
  95.6 &
  52.2 &
  81.6 &
  95.8 \\
 &
  AP &
  44.1 &
  14.9 &
  51.6 &
  42.3 &
  {\underline{61.4}} &
  \textbf{61.6} &
  50.8 &
  52.8 &
  4.8 &
  24.0 &
  52.3 \\
 &
  Max-F1 &
  49.2 &
  23.5 &
  56.7 &
  49.3 &
  {\underline{61.3}} &
  \textbf{64.6} &
  54.2 &
  56.8 &
  9.0 &
  28.5 &
  54.9 \\
 &
  AUPRO &
  78.8 &
  53.2 &
  {\underline{90.3}} &
  86.1 &
  87.3 &
  \textbf{94.4} &
  79.8 &
  86.2 &
  17.7 &
  51.8 &
  76.2 \\ \bottomrule[1.5pt]
\end{tabular}
}
\end{table*}
\begin{figure*}[h!]
\centering\includegraphics[width=\linewidth]{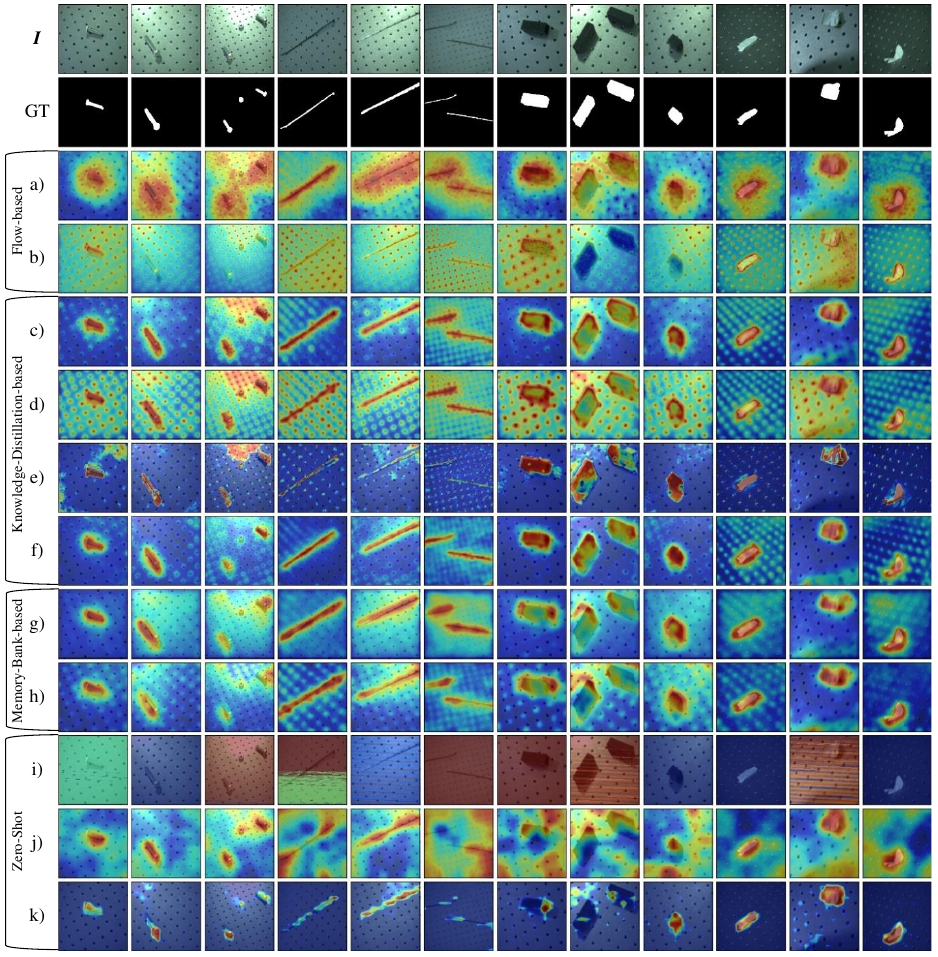}
\caption{Visualization of prediction results using the selected methods. 
Every three columns represent a category, with the first column representing high-quality sample, the second column representing samples with uneven illumination, and the third column representing samples with blurry. The first row is the original images collected, while the second row is the pixel-wise mask. Subsequent rows depict various methods: a) Cflow b) PyramidFlow c) RD d) RD++ e) DSR f) CDO g) Patchcore h) GCPF i) SAA j) WinCLIP k) APRIL-GAN. }
\label{fig:vis}
\end{figure*}
\subsection{Experiment Settings}
\subsubsection{Comparison Methods}

Representative methods, including both unsupervised anomaly detection and zero-shot anomaly detection methods, are selected for evaluation and analysis. Considering that real abnormal samples are difficult to obtain, unsupervised anomaly detection methods have emerged as a research hotspot~\cite{liu2024deep}. These methods are trained solely on normal samples and are capable of both image-level and pixel-level anomaly detection.

Unsupervised methods encompass multiple research branches. Specifically, CFlow~\cite{cflow} and PyramidFlow~\cite{PyramidFlow} are flow-based methods that model the probability distribution from pre-trained features and locate anomalies by the distance in the function variable space. RD~\cite{RD}, RD++~\cite{RD++}, DSR~\cite{DSR}, and CDO~\cite{cdo} are knowledge-distillation-based methods that minimize the feature distance between normal samples in a teacher-student model, allowing anomaly detection through significant feature differences in the model. PatchCore~\cite{patchcore} and GCPF~\cite{GCPF} are memory-bank-based methods that calculate the feature distribution of normal samples and determine anomalies by measuring the distance between the tested image and the normal sample distribution.

In addition, some zero-shot anomaly detection methods~\cite{winclip,saa,APRIL-GAN} utilize the general knowledge of foundation models, achieving promising performance without training with any normal or abnormal samples. WinCLIP~\cite{winclip} and APRIL-GAN~\cite{APRIL-GAN} employ a pre-trained image encoder to extract rich image features and design text descriptions to prompt anomaly detection. SAA~\cite{saa} combines two off-the-shelf foundation models to segment any anomalies in a user-friendly manner.


\subsubsection{Evaluation Metrics}
Following the mainstream setting as in MVTec-AD~\cite{mvtec}, we employ several key metrics to evaluate the performance of anomaly detection methods, namely the Area Under the Receiver Operating Characteristic curve (AUROC), Maximal F1 score (Max-F1), Average Precision (AP), and Area Under Per Region Overlap (AUPRO). The first three indicators (AUROC, Max-F1, AP) are utilized for both image-level and pixel-level anomaly detection evaluation, and AUPRO is exclusively employed for pixel-level anomaly detection evaluation.

\subsubsection{Implementation Details}
Experiments are conducted using publicly available model source codes. The input image is resized to a resolution of $224\times224$. Memory-bank-based methods, SAA and WinCLIP require no training. For the other methods, the epoch demonstrating the best performance is selected. 
It is worth noting that training of APRIL-GAN is conducted on abnormal data from MVTec-AD. Models training and inference are performed by using PyTorch version 1.7 on a single GPU RTX A6000.

\subsection{Experiment Results}


Table~\ref{table:image-level} presents the image-level anomaly detection performance of various methods on the \dataset{} dataset. RD++ and WinCLIP achieve the best performance, with WinCLIP reaching 98.6\% AUROC and 99.7\% AP, and RD++ achieving 98.6\% Max-F1. Table~\ref{table:pixel-level} presents the pixel-level results of different methods on the \dataset{} dataset. Unlike the image-level results, RD++ and WinCLIP exhibit a significant decrease in pixel-level performance. In contrast, CDO and RD achieved higher pixel-level performance but still failed to meet practical requirements. Fig.~\ref{fig:vis} visualizes the prediction results of the evaluated methods.

It is evident that flow-based methods generate many false positives. Knowledge-distillation-based methods and memory-bank-based methods can detect foreign objects in most cases but are also sensitive to background holes. Although WinCLIP shows good image-level performance, it is almost ineffective in pixel-level anomaly detection. The performance of all zero-shot methods is lower than that of unsupervised methods, as they lack specific knowledge of the target categories.

\subsection{Experiment Analysis}

\subsubsection{Free Views}
The impact of free views mainly lies in misaligned foreign objects and misaligned background holes. Visualization results indicate that while most methods accurately identify foreign objects as anomalies, nearly all struggle to completely avoid interference from background holes. The flow-based methods generate high anomaly scores around the foreign objects for their no-compact normal distribution estimation. 
The knowledge-distillation-based methods and the memory-bank-based methods also face challenges in aligning background features due to the randomness of background holes.

\subsubsection{Uneven Illuminations}
The bright spots generated by uneven illuminations indeed induce the model to output incorrect results, as shown in the second column of each category in Fig.~\ref{fig:vis}. Irregular light spots still pose obstacles to the probability estimation of flow-based methods. Because of the significant differences between the spot and normal images and it is hard to minimize the distance between the spot and the normal images, the knowledge-distillation-based methods are also somewhat ineffective. An exception is CDO that increases the distribution distance between normal and abnormal samples by generating anomalies, making the model more sensitive to foreign objects than uneven illuminations. The memory-bank-based methods exhibit stronger robustness against uneven illuminations. Patchcore employs mid-level feature clustering and coreset sampling to improve generalization. GCPF utilizes more cluster centers, thereby more effectively addressing illumination effects.

\subsubsection{Blurry Collections}
Blurry images, as shown in the third column of each category in Fig.~\ref{fig:vis}, result in lower anomaly scores compared to high-quality images in the first column of each category in Fig.~\ref{fig:vis}. The presence of blurry noise increases the model's uncertainty in anomaly detection. Above all anomaly detection methods utilize image encoders to extract and utilize the semantic information of images, which typically has strong robustness against low dimensional noises (blurs).

Despite zero-shot methods scoring lower in quantitative evaluations compared to unsupervised methods, visualization results highlight the potential robustness of APRIL-GAN. While APRIL-GAN does not precisely detect the boundaries of foreign objects, it significantly identifies parts of these objects. Crucially, APRIL-GAN is minimally affected by light spots and blur. This can be attributed to the foundation model's powerful general knowledge and accurate text descriptions. During pre-training on the MVTec-AD dataset, the foundation model is better able to concentrate on image anomalies. With the prompting of text and the strong feature extraction ability of the foundation model, APRIL-GAN can more effectively distinguish content unrelated to the task (uneven illuminations, background holes, blurs, etc.). 
Overall, we find that the memory-bank-based methods have better robustness for views and illuminations. In addition, the artificially synthesized anomalies also help to enhance the robustness of the model. Effectively leveraging the general knowledge of foundation models presents a promising avenue for enhancing anomaly detection robustness.




\section{CONCLUSION}\label{sec:conclusion}
To simulate actual industrial detection conditions to evaluate the robustness of anomaly detection methods, we build the \dataset{} dataset. \dataset{} generates anomalies by placing foreign objects on the work platform. The dataset comprises four types of foreign object: Bolt, Cable\_tie, Sponge and Tape. By varying the views of collection, the uniformity of environmental illuminations, and the collection distances or stability, we obtain a large dataset of misaligned, low-quality industrial images, both normal and abnormal.
Extensive experiments have been conducted to evaluate the robustness of both unsupervised and zero-shot methods, bringing the following conclusions:
\begin{itemize}
\item The non alignment of background features caused by free views almost leads to a large number of false positives in all anomaly detection methods, and the light spots caused by uneven illuminations are easily identified as anomalies. Blurs mainly brings uncertainty to the abnormal scores. Current anomaly detection methods still have a certain distance from complex detection environments.
\item The memory-bank-based methods have better robustness by extracting and compressing feature distributions. In addition, utilizing synthesized anomalies can enhance the model's understanding of normal data and weaken the interference of illuminations.
\item Zero-shot methods, leveraging foundation models, achieve a generalized representation of image features. Text prompts enable a focused analysis on foreign objects, reducing the impact of environmental variations. Combining the advantages of the foundation model, the robustness of anomaly detection methods can be further enhanced.
\end{itemize}






\section*{ACKNOWLEDGMENT}

The computation is completed in the HPC Platform of Huazhong University of Science and Technology.



\begin{thebibliography}{10}

\bibitem{cao2024survey}
Yunkang Cao, Xiaohao Xu, Jiangning Zhang, Yuqi Cheng, Xiaonan Huang, Guansong Pang, and Weiming Shen.
\newblock A survey on visual anomaly detection: Challenge, approach, and prospect.
\newblock {\em arXiv preprint arXiv:2401.16402}, 2024.

\bibitem{wang2024real}
Chengjie Wang, Wenbing Zhu, Bin-Bin Gao, Zhenye Gan, Jianning Zhang, Zhihao Gu, Shuguang Qian, Mingang Chen, and Lizhuang Ma.
\newblock Real-iad: A real-world multi-view dataset for benchmarking versatile industrial anomaly detection.
\newblock {\em arXiv preprint arXiv:2403.12580}, 2024.

\bibitem{VisA}
Yang Zou, Jongheon Jeong, Latha Pemula, Dongqing Zhang, and Onkar Dabeer.
\newblock Spot-the-difference self-supervised pre-training for anomaly detection and segmentation.
\newblock In {\em European Conference on Computer Vision}, pages 392--408. Springer, 2022.

\bibitem{mvtec}
Paul Bergmann, Michael Fauser, David Sattlegger, and Carsten Steger.
\newblock Mvtec ad--a comprehensive real-world dataset for unsupervised anomaly detection.
\newblock In {\em Proceedings of the IEEE/CVF conference on computer vision and pattern recognition}, pages 9592--9600, 2019.

\bibitem{zhang2023exploring}
Jiangning Zhang, Xuhai Chen, Yabiao Wang, Chengjie Wang, Yong Liu, Xiangtai Li, Ming-Hsuan Yang, and Dacheng Tao.
\newblock Exploring plain vit reconstruction for multi-class unsupervised anomaly detection.
\newblock {\em arXiv preprint arXiv:2312.07495}, 2023.

\bibitem{yao2023scalable}
Haiming Yao, Wei Luo, Jianan Lou, Wenyong Yu, Xiaotian Zhang, Zhenfeng Qiang, and Hui Shi.
\newblock Scalable industrial visual anomaly detection with partial semantics aggregation vision transformer.
\newblock {\em IEEE Transactions on Instrumentation and Measurement}, 2023.

\bibitem{xie2023iad}
Guoyang Xie, Jinbao Wang, Jiaqi Liu, Jiayi Lyu, Yong Liu, Chengjie Wang, Feng Zheng, and Yaochu Jin.
\newblock Im-iad: Industrial image anomaly detection benchmark in manufacturing.
\newblock {\em arXiv preprint arXiv:2301.13359}, 2023.

\bibitem{akcay2022anomalib}
Samet Akcay, Dick Ameln, Ashwin Vaidya, Barath Lakshmanan, Nilesh Ahuja, and Utku Genc.
\newblock Anomalib: A deep learning library for anomaly detection.
\newblock In {\em 2022 IEEE International Conference on Image Processing (ICIP)}, pages 1706--1710. IEEE, 2022.

\bibitem{cheng2022novel}
Yu-qi Cheng, Wen-long Li, Cheng Jiang, Dong-fang Wang, Jin-cheng Mao, and Wei Xu.
\newblock A novel point cloud simplification method using local conditional information.
\newblock {\em Measurement Science and Technology}, 33(12):125203, 2022.

\bibitem{cheng2021novel}
Yu-qi Cheng, Wen-long Li, Cheng Jiang, Gang Wang, Wei Xu, and Qing-yu Peng.
\newblock A novel cooling hole inspection method for turbine blade using 3d reconstruction of stereo vision.
\newblock {\em Measurement Science and Technology}, 33(1):015018, 2021.

\bibitem{MPDD}
Stepan Jezek, Martin Jonak, Radim Burget, Pavel Dvorak, and Milos Skotak.
\newblock Deep learning-based defect detection of metal parts: evaluating current methods in complex conditions.
\newblock In {\em 2021 13th International congress on ultra modern telecommunications and control systems and workshops (ICUMT)}, pages 66--71. IEEE, 2021.

\bibitem{eyecandies}
Luca Bonfiglioli, Marco Toschi, Davide Silvestri, Nicola Fioraio, and Daniele De~Gregorio.
\newblock The eyecandies dataset for unsupervised multimodal anomaly detection and localization.
\newblock In {\em Proceedings of the Asian Conference on Computer Vision}, pages 3586--3602, 2022.

\bibitem{cflow}
Denis Gudovskiy, Shun Ishizaka, and Kazuki Kozuka.
\newblock Cflow-ad: Real-time unsupervised anomaly detection with localization via conditional normalizing flows.
\newblock In {\em Proceedings of the IEEE/CVF Winter Conference on Applications of Computer Vision}, pages 98--107, 2022.

\bibitem{PyramidFlow}
Jiarui Lei, Xiaobo Hu, Yue Wang, and Dong Liu.
\newblock Pyramidflow: High-resolution defect contrastive localization using pyramid normalizing flow.
\newblock In {\em Proceedings of the IEEE/CVF Conference on Computer Vision and Pattern Recognition}, pages 14143--14152, 2023.

\bibitem{RD}
Hanqiu Deng and Xingyu Li.
\newblock Anomaly detection via reverse distillation from one-class embedding.
\newblock In {\em Proceedings of the IEEE/CVF Conference on Computer Vision and Pattern Recognition}, pages 9737--9746, 2022.

\bibitem{RD++}
Tran~Dinh Tien, Anh~Tuan Nguyen, Nguyen~Hoang Tran, Ta~Duc Huy, Soan Duong, Chanh D~Tr Nguyen, and Steven~QH Truong.
\newblock Revisiting reverse distillation for anomaly detection.
\newblock In {\em Proceedings of the IEEE/CVF Conference on Computer Vision and Pattern Recognition}, pages 24511--24520, 2023.

\bibitem{DSR}
Vitjan Zavrtanik, Matej Kristan, and Danijel Sko{\v{c}}aj.
\newblock Dsr--a dual subspace re-projection network for surface anomaly detection.
\newblock In {\em European conference on computer vision}, pages 539--554. Springer, 2022.

\bibitem{cdo}
Yunkang Cao, Xiaohao Xu, Zhaoge Liu, and Weiming Shen.
\newblock Collaborative discrepancy optimization for reliable image anomaly localization.
\newblock {\em {IEEE} Transactions on Industrial Informatics}, pages 1--10, 2023.

\bibitem{patchcore}
Karsten Roth, Latha Pemula, Joaquin Zepeda, Bernhard Sch{\"o}lkopf, Thomas Brox, and Peter Gehler.
\newblock Towards total recall in industrial anomaly detection.
\newblock In {\em Proceedings of the IEEE/CVF Conference on Computer Vision and Pattern Recognition}, pages 14318--14328, 2022.

\bibitem{GCPF}
Qian Wan, Liang Gao, Xinyu Li, and Long Wen.
\newblock Industrial image anomaly localization based on gaussian clustering of pretrained feature.
\newblock {\em IEEE Transactions on Industrial Electronics}, 69(6):6182--6192, 2021.

\bibitem{saa}
Yunkang Cao, Xiaohao Xu, Chen Sun, Yuqi Cheng, Zongwei Du, Liang Gao, and Weiming Shen.
\newblock Segment any anomaly without training via hybrid prompt regularization.
\newblock {\em arXiv preprint arXiv:2305.10724}, 2023.

\bibitem{winclip}
Jongheon Jeong, Yang Zou, Taewan Kim, Dongqing Zhang, Avinash Ravichandran, and Onkar Dabeer.
\newblock Winclip: Zero-/few-shot anomaly classification and segmentation.
\newblock In {\em Proceedings of the IEEE/CVF Conference on Computer Vision and Pattern Recognition}, pages 19606--19616, 2023.

\bibitem{APRIL-GAN}
Xuhai Chen, Yue Han, and Jiangning Zhang.
\newblock A zero-/few-shot anomaly classification and segmentation method for cvpr 2023 vand workshop challenge tracks 1\&2: 1st place on zero-shot ad and 4th place on few-shot ad.
\newblock {\em arXiv preprint arXiv:2305.17382}, 2023.

\bibitem{liu2024deep}
Jiaqi Liu, Guoyang Xie, Jinbao Wang, Shangnian Li, Chengjie Wang, Feng Zheng, and Yaochu Jin.
\newblock Deep industrial image anomaly detection: A survey.
\newblock {\em Machine Intelligence Research}, 21(1):104--135, 2024.

\end{thebibliography}


\end{document}